%% file: main_text.tex
\documentclass{article}

\pdfoutput=1
\usepackage{Others/PRIMEarxiv}

\usepackage[utf8]{inputenc} 
\usepackage[T1]{fontenc}    
\usepackage{url}            
\usepackage{booktabs}       
\usepackage{amsfonts}       
\usepackage{amsmath}
\usepackage{nicefrac}       
\usepackage{microtype}      
\usepackage{lipsum}
\usepackage{fancyhdr}       
\usepackage{graphicx}       
\graphicspath{{Figures/}}     
\usepackage{chngcntr}

\usepackage{biblatex}
\usepackage[pdftex]{hyperref}

\input{Sections/00_Title}

\begin{document}
\maketitle

\input{Sections/01_Abstract}
\input{Sections/02_Introduction}
\input{Sections/03_Theory}
\input{Sections/04_Examples}
\input{Sections/05_Discussion}
\input{Sections/06_Conclusion}
\input{Sections/07_Acknowledgement}

\printbibliography

\end{document}

%% file: Sections/00_Title.tex
\title{Laplace HypoPINN: Physics-Informed Neural Network for hypocenter localization and its predictive uncertainty}

\author{
  Muhammad Izzatullah\thanks{\textbf{Corresponding author:} \texttt{muhammad.izzatullah@kaust.edu.sa}} , Isa Eren Yildirim, Tariq Alkhalifah \\
  King Abdulllah University of Science and Technology (KAUST) \\
  \AND
  Umair Bin Waheed \\
  King Fahd University of Petroleum and Minerals (KFUPM) \\
}

%% file: Sections/01_Abstract.tex
\begin{abstract}
Several techniques have been proposed over the years for automatic hypocenter localization. While those techniques have pros and cons that trade-off computational efficiency and the susceptibility of getting trapped in local minima, an alternate approach is needed that allows robust localization performance and holds the potential to make the elusive goal of real-time microseismic monitoring possible. Physics-informed neural networks (PINNs) have appeared on the scene as a flexible and versatile framework for solving partial differential equations (PDEs) along with the associated initial or boundary conditions. We develop \emph{HypoPINN} -- a PINN-based inversion framework for hypocenter localization and introduce an approximate Bayesian framework for estimating its predictive uncertainties. This work focuses on predicting the hypocenter locations using HypoPINN and investigates the propagation of uncertainties from the random realizations of HypoPINN's weights and biases using the Laplace approximation. We train HypoPINN to obtain the optimized weights for predicting hypocenter location. Next, we approximate the covariance matrix at the optimized HypoPINN's weights for posterior sampling with the Laplace approximation. The posterior samples represent various realizations of HypoPINN's weights. Finally, we predict the locations of the hypocenter associated with those weights' realizations to investigate the uncertainty propagation that comes from those realisations. We demonstrate the features of this methodology through several numerical examples, including using the Otway velocity model based on the Otway project in Australia. 
\end{abstract}

\keywords{Deep learning, Approximate probabilistic inference, Simulation, Unsupervised learning, Neural networks}

%% file: Sections/02_Introduction.tex
\section{Introduction}
In recent years, seismicity induced by anthropogenic activities including underground mining, geothermal exploitation, hydrofracturing, CO$_2$ geologic sequestration, and hydrocarbon production has resulted in a sharp increase in the number of earthquakes observed in historically quiet tectonic areas. In addition to causing considerable economic losses, such events are increasingly becoming a threat to public safety. A traffic light system (TLS) is typically implemented to manage and mitigate the associated hazard by reducing or suspending operations in case of observed seismicity beyond preset thresholds \cite{mignan2017induced}. The success of a TLS protocol relies on real-time capabilities of detecting and locating these events.

Several techniques have been proposed over the years for automatic hypocenter localization (see~\cite{foulger2018global} for an overview). Early methods relied on picked arrival times and estimated the unknown source location coordinates by using time difference as the objective function~\cite{lienert1986hypocenter,pugh2016bayesian}. These methods construct different objective functions to obtain absolute~\cite{bancroft2010sensitivity} or relative~\cite{waldhauser2000double} locations. More recent approaches use full seismic waveforms to image the source location using a migration-type method~\cite{zhang2015genetic}. While these approaches have pros and cons that trade-off computational efficiency and the susceptibility of getting trapped in local minima, an alternate approach is needed that allows robust localization performance and holds the potential to make the elusive goal of real-time microseismic monitoring possible.

The confluence of ultrafast computers, rapid advancements in machine learning algorithms, and increasing availability of large datasets place seismology at the threshold of dramatic progress. Therefore, it is no surprise that several localization approaches have recently been proposed to harness the potential of supervised machine learning. These methods typically train a convolutional neural network (CNN) using historical or synthetically generated datasets~\cite{perol2018convolutional,wang2021data}. Once the CNN model is trained, it can be used to infer locations in real-time. Nevertheless, these methods typically require a huge amount of training data that may not be readily available. More importantly, due to the black-box nature of these models, it is not easy to gain insights into the features learned by the model.

Physics-informed neural networks (PINNs) have appeared on the scene as a flexible and versatile framework for solving partial differential equations (PDEs), along with any initial or boundary conditions \cite{Raissi2019}. Recently, researchers have explored the potential of PINNs as a fast travel time modeling engine for hypocenter localization~\cite{grubas2021localization,smith2022hyposvi}. For other seismic applications, \cite{Song2021} and \cite{Alkhalifah2021} used PINNs as a solver for the wave equation, including inverting for the velocity model. \cite{Waheed2021} utilized it to provide a framework to solve the eikonal equation and \cite{Waheed2021tomo} extended the framework for seismic tomography. An important component of these solutions, especially when using the data as a boundary condition, is our confidence in their accuracy. There has been little study of PINN accuracy as an inversion tool. Neural networks naturally embed stochasticity through the random realization of weights and biases, thus, propagating uncertainties into its predictive solutions \cite{Yang2021}. 

Our contributions in this work are two-fold: We develop a direct inversion framework for hypocenter localization using PINNs and introduce an approximate Bayesian framework for estimating its predictive uncertainties. Given picked arrival times for an event, we train a PINN model by minimizing a loss function formed by the misfit of observed and predicted travel times and the residual of the eikonal equation and obtain a travel time map for the entire computational domain. The hypocenter locations are then obtained by finding the location of the minimum travel time, representing the focusing location within the domain. As a result, we refer to this inversion framework as \emph{HypoPINN}. We also investigate the propagation of uncertainties from the random realizations of HypoPINN's weights and biases using the Laplace approximation\cite{Ritter2018, Daxberger2021}.  

The Laplace approximation is arguably the simplest family of approximations for the intractable posteriors of deep neural networks. We approximate the covariance matrix at the optimized HypoPINN's weights for posterior sampling with the Laplace approximation. The posterior samples represent various realizations of HypoPINN's weights. Finally, we predict the locations of the hypocenter associated with those weights' realizations to investigate the uncertainty propagation that comes from the weights' realizations. The uncertainties estimation from this approach is called predictive uncertainty or, simply, forward modeling uncertainty in the context of HypoPINN.

The outline of the rest of the paper is as follows. First, we introduce the problem formulation of estimating hypocenter locations through PINN, which is the basis of HypoPINN and its predictive uncertainty studies. Next, we discuss the theoretical framework of the proposed approximate Bayesian framework for estimating predictive uncertainties in HypoPINN based on Laplace approximation. Then, we demonstrate the features of this methodology through several numerical examples, including using the Otway velocity model. Finally, we discuss the limitations and possible improvements of the proposed method before concluding the study.

%% file: Sections/03_Theory.tex
\section{Theoretical Framework}
We focus on developing a direct inversion framework for hypocenter localizations directly using PINNs by solving the eikonal equation and investigating its predictive uncertainties. Several works have shown promising results in solving the eikonal equation through PINN \cite{Waheed2021, waheed2021holistic, waheed2021pinntomo, Waheed2021tomo} by leveraging the factorization idea of eikonal equation \cite{fomel2009factored}. However, this approach could not be extended to our work because the factorized eikonal equation requires the source locations to be known. Thus, we utilize the original eikonal equation together with the traveltime observations recorded at the surface to predict a traveltime map and by taking its minimum, we can directly localize the hypocenter locations.

\begin{figure*}[!htb]
  \centering
  \includegraphics[width=\textwidth]{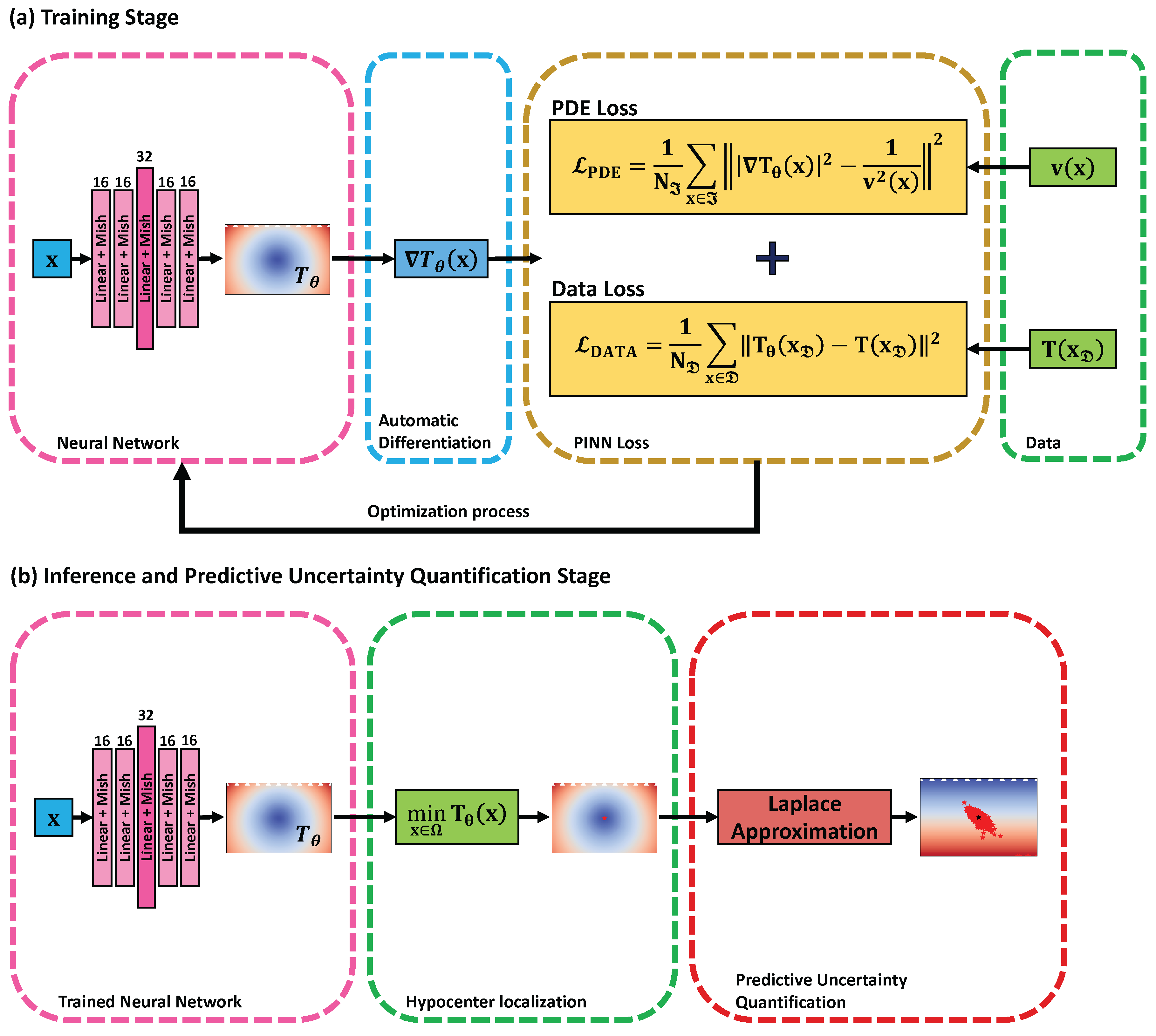}
  \caption{HypoPINN workflow for hypocenter localization and predictive uncertainty quantification: (a) Training stage, (b) inference and predictive uncertainty quantification stage.}
  \label{fig:workflow}
\end{figure*}

In this section, we first introduce the eikonal equation and its physics-informed neural network (PINN) representation for the hypocenter localization problem. This is followed by a brief overview of PINN in the Bayesian framework and the Laplace approximation as an approximator for the intractable posterior distribution in quantifying its predictive uncertainty. Finally, putting these pieces together, we present the proposed algorithm for solving the hypocenter localization problem through PINN, which we refer to as \emph{HypoPINN}. The full HypoPINN's workflow is illustrated in Fig. \ref{fig:workflow}.

\subsection{HypoPINN for hypocenter localisation}
Here, we outline the HypoPINN formulation for hypocenter localization through the eikonal equation. The eikonal equation is a non-linear, first-order, hyperbolic PDE of the form:
\begin{equation}\label{eq:1}
    |\nabla \mathbf{T}(\mathbf{x})|^{2} = \frac{1}{v^{2}(\mathbf{x})}, \quad \forall \mathbf{x} \in \Omega
\end{equation}
where $\Omega$ is a domain in $\mathrm{R}^{d}$ with $d$ as the space dimension, $\mathbf{T}(\mathbf{x})$ is the travel time from the point-source $\mathbf{x}_{s}$ to any point $\mathbf{x}$, $v(\mathbf{x})$ is the velocity defined in $\Omega$, and $\nabla$ denotes the spatial differential operator. HypoPINN can be formulated as an optimization problem for the learnable PINN's parameters $\mathbf{\theta}$ in approximating the traveltimes and estimating the hypocenter. The loss function for solving HypoPINN can be constructed using a mean-squared error (MSE) loss as:
\begin{equation}\label{eq:2}
\begin{split}
\mathcal{L}(\mathbf{\theta}) & = \frac{1}{N_{\mathcal{I}}} \sum_{\mathbf{x}^{\ast} \in \mathcal{I}} \||\nabla \mathbf{T}_{\mathbf{\theta}}(\mathbf{x}^{\ast})|^{2} - \frac{1}{v^{2}(\mathbf{x}^{\ast})}\|^{2}\\ 
& + \frac{1}{N_{\mathcal{D}}} \sum_{\hat{\mathbf{x}} \in \mathcal{D}} \|\mathbf{T}_{\mathbf{\theta}}(\hat{\mathbf{x}}) - \mathbf{T}(\hat{\mathbf{x}})\|^{2},
\end{split}
\end{equation}
where $\mathbf{T}_{\mathbf{\theta}}(\mathbf{x})$ represents the neural networks for the eikonal solution $\mathbf{T}(\mathbf{x})$. The first term on the right side of equation (\ref{eq:2}) imposes the validity of the eikonal equation as in (\ref{eq:1}) on a given set of training points $\mathbf{x}^{\ast} \in \mathcal{I}$, with $N_{\mathcal{I}}$ as the number of sampling points. The second term acts as data loss on a given set of travel time data at the receiver locations $\hat{\mathbf{x}} \in \mathcal{D}$, with $N_{\mathcal{D}}$ representing the number of receivers. We minimize the loss in (\ref{eq:2}) to obtain a good approximation of the eikonal solution and hypocenter location. Once the network is trained, the solution (microseismic hypocenter location) is basically the minimum of the computed travel time function, evaluated within the computational domain:
\begin{equation}
\label{eqn:source_location}
\mathbf{T}_{\mathbf{\theta}^{\ast}_{\min}} = \min_{\hat{\mathbf{x}} \in \Omega}\{\mathbf{T}_{\mathbf{\theta}^{\ast}}(\hat{\mathbf{x}})\} .
\end{equation}

\subsection{Laplace Approximation for Bayesian PINNs}
Bayesian framework for PINNs can be formulated through \emph{unnormalized} Bayes' Theorem \cite{Yang2021} as
\begin{equation}\label{eq:8}
    \mathbf{p}(\mathbf{\theta}|\mathbf{D}) \propto \mathbf{p}(\mathbf{D}|\mathbf{\theta})\mathbf{p}(\mathbf{\theta}) \approx \exp{\Big(-\mathcal{L}(\mathbf{D}; \mathbf{\theta})\Big)},
\end{equation}
where $\mathbf{\theta}$ denotes the learnable PINN's parameters and $\mathbf{D}$ represents the dataset associated with PINN's training, e.g., observed data collected by seismic stations. The last term in equation (\ref{eq:8}) is known as the \emph{Gibbs distribution}. We can transform the \emph{Gibbs distribution} in (\ref{eq:8}) into a representation similar to the loss function in a deterministic setting by reformulating it in the \emph{log-posterior} as follows:
\begin{equation}\label{eq:3}
    \log \mathbf{p}(\mathbf{\theta}|\mathbf{D}) \propto -\log \mathbf{p}(\mathbf{D}|\mathbf{\theta}) - \log \mathbf{p}(\mathbf{\theta}) \approx \mathcal{L}(\mathbf{D}; \mathbf{\theta}).
\end{equation}
By minimizing (\ref{eq:3}), we obtain the \emph{Maximum-A-Posteriori (MAP)} solution that we consider as the centre of our Laplace approximation. The Laplace approximation uses a second-order expansion (Taylor expansion) of $\mathcal{L}(\mathbf{D}; \mathbf{\theta})$ around $\mathbf{\theta}_{MAP}$ to approximate $\mathbf{p}(\mathbf{\theta}|\mathbf{D})$. We consider
\begin{equation}\label{eq:4}
    \begin{split}
        \mathcal{L}(\mathbf{D}; \mathbf{\theta}) &\approx  \mathcal{L}(\mathbf{D}; \mathbf{\theta}_{MAP}) \\
        &+ \frac{1}{2}(\mathbf{\theta} - \mathbf{\theta}_{MAP})^{T}\Big(\nabla^{2}_{\mathbf{\theta}}\mathcal{L}(\mathbf{D}; \mathbf{\theta}_{MAP}) \Big)(\mathbf{\theta} - \mathbf{\theta}_{MAP}),
    \end{split}
\end{equation}
and identify the Laplace approximation for $\mathbf{p}(\mathbf{\theta}|\mathbf{D})$ as
\begin{equation}\label{eq:5}
    \mathbf{p}(\mathbf{\theta}|\mathbf{D}) \approx \mathcal{N}(\mathbf{\theta}_{MAP}, \Sigma), \quad \text{with} \quad \Sigma = - \Big(\nabla^{2}_{\mathbf{\theta}}\mathcal{L}(\mathbf{D}; \mathbf{\theta}_{MAP}) \Big)^{-1}. 
\end{equation}
Note that a naive implementation of the covariance matrix in (\ref{eq:5}) is infeasible, and it scales quadratically with the number of learnable PINN's parameters, $\mathbf{\theta}$. This work focuses on the diagonal approximation for the covariance matrix. Interested readers may refer to \cite{Ritter2018} for a detailed review on the scalable Laplace approximation for Bayesian neural networks. The diagonal approximation of the covariance matrix based on the Fisher information matrix $\mathbf{F}$ can be computed efficiently using automatic differentiation. It is simply the expectation of the squared gradients with respect to the network parameters $\mathbf{\theta}$:
\begin{equation*}
\begin{split}
    \mathbf{H} &\approx \text{diag}(\mathbf{F}) \\
    &= \text{diag}\Big(\mathrm{E}\Big[\nabla_{\mathbf{\theta}}\mathcal{L}(\mathbf{D}; \mathbf{\theta})\nabla_{\mathbf{\theta}}\mathcal{L}(\mathbf{D}; \mathbf{\theta})^{T} \Big] \Big) \\
    &= \text{diag}\Big(\mathrm{E}\Big[(\nabla_{\mathbf{\theta}}\mathcal{L}(\mathbf{D}; \mathbf{\theta}))^{2} \Big] \Big),
\end{split}
\end{equation*}
where "$\text{diag}$" extracts the diagonal of a matrix. Note that, even if the expansion in (\ref{eq:4}) is accurate, this approximation will, unfortunately, place probability mass in low probability regions of the true posterior if some of the PINN's parameters $\mathbf{\theta}$ exhibit high covariance. Despite the fact, it has been used successfully in neural network weights pruning and transfer learning \cite{Kirkpatrick2017}. Based on the diagonal approximation, we can approximate our covariance by
\begin{equation}\label{eq:7}
    \Sigma \approx \mathbf{H}^{-1} = \frac{1}{\text{diag}(\mathbf{F})}.
\end{equation}
To apply the Laplace approximation for uncertainty estimation, we first minimize (\ref{eq:3}) to obtain the $\mathbf{\theta}_{MAP}$. We can transform the loss function in (\ref{eq:2}) into Bayesian framework as in (\ref{eq:3}) by reformulating it in the \emph{log-posterior} form with a chosen log-prior distribution $\log \mathbf{p}(\mathbf{\theta})$ that commonly acts as a regularizer in a deterministic setting. Next, we approximate the covariance matrix at $\mathbf{\theta}_{MAP}$ and construct the Laplace approximation of the posterior distribution as in (\ref{eq:5}). The posterior samples represent various realizations of PINN's weights, $\mathbf{\theta}$. Finally, we predict the solutions associated with those weights' realizations $\mathbf{\theta}$ to investigate the uncertainty propagation that comes from those realizations.

%% file: Sections/04_Examples.tex
\section{Numerical Examples}
This section demonstrates the proposed methodology on numerical examples with a vertically varying velocity model of $2 \times 3$ km$^2$ that varies with depth and the Otway velocity model. In all numerical examples, we specially crafted a neural network architecture with an expansion-contraction design for HypoPINN in predicting the eikonal solution and localizing the hypocenter locations. The neural network contains $5$ fully connected hidden layers and $16$ neurons per layer except for the third layer with $32$ neurons. Intuitively, by doubling the neurons in the middle layer provides the network with more degrees of freedom in efficiently transforming the inputs (i.e., spatial coordinates) which belong to the euclidean space into polar coordinates space which represents the traveltimes map. We consider \emph{Mish} \cite{Misra2019mish} as the activation function and Adam optimiser \cite{kingma2014adam} as the optimisation algorithm. The network architecture is illustrated in Fig. \ref{fig:workflow}(a).

\subsection{The vertically varying velocity model}
\begin{figure*}[!htb]
  \centering
  \includegraphics[width=\textwidth]{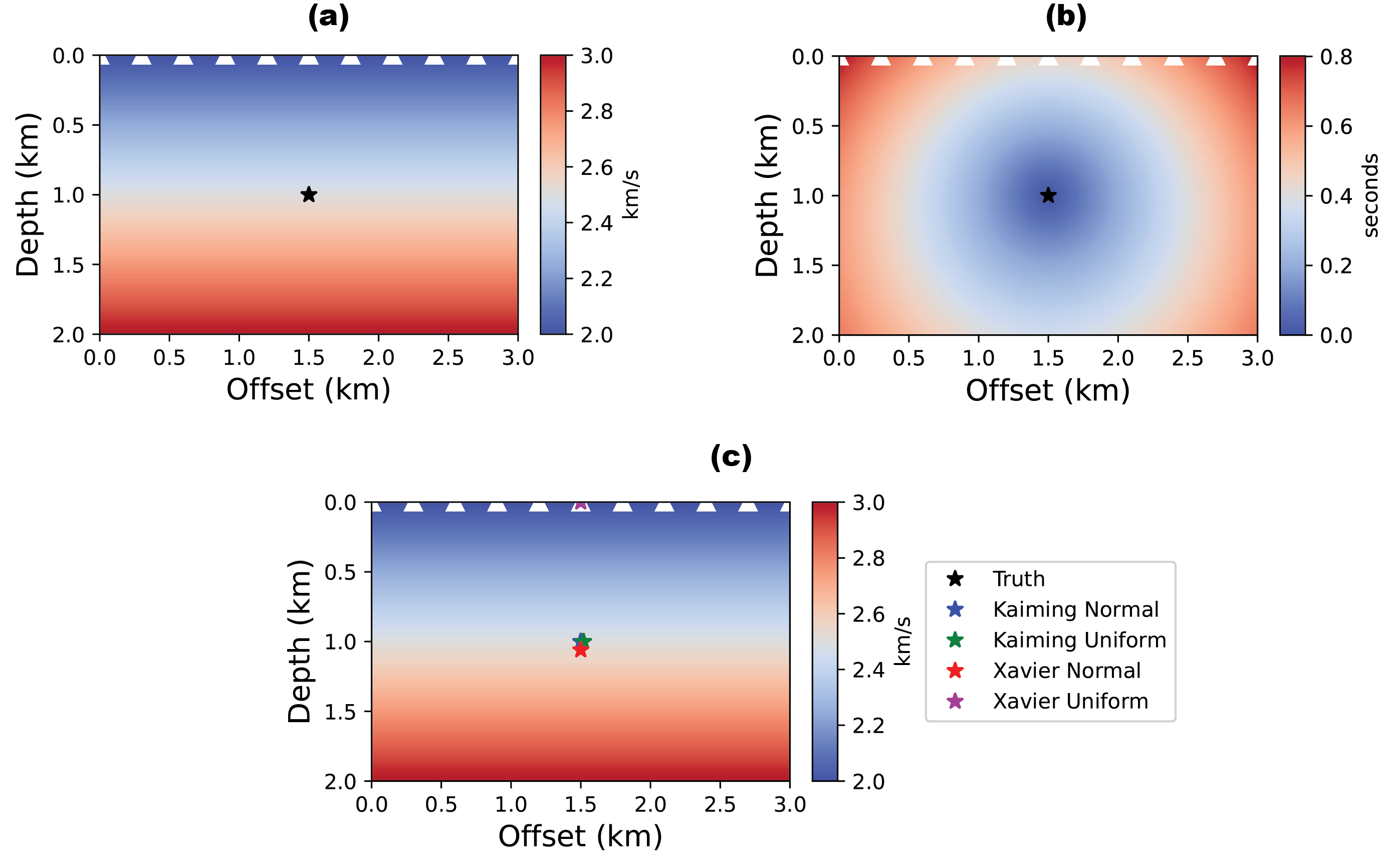}
  \caption{Random weights initializations example: (a) vertically varying velocity model, (b) true eikonal solution, (c) estimated hypocenter locations using HypoPINN for four different weights initializations sampled from the selected prior distributions. The black star and white triangles denote the true source and receiver locations.}
  \label{fig:1}
\end{figure*}

We consider a vertically varying velocity model of $2 \times 3$ km$^2$ that varies with depth. The velocity at zero depth is $2$ km/s, and it increases linearly with a gradient of $0.5 s^{-1}$. We place a point-source at ($1$ km, $1.5$ km) representing the microseismic event. The model is illustrated in Fig. \ref{fig:1}(a) with the black star depicting the point-source location. The model is discretized on a $101 \times 151$ grid with a grid spacing of $20$ m along both axes. The true traveltime map is computed analytically \cite{slotnick1959lessons}. 

First, we investigate the effect of random initial weights initialization from four different prior distributions as a motivation towards understanding the hypocenter location prediction performance and the predictive uncertainties of HypoPINN. We consider the modified normal and uniform distributions introduced by \cite{glorot2010} and \cite{kaiminghe2015} for the initial weights. We randomly sample $2500$ points in the domain and collect the travel time value at $11$ receivers on the surface as data for training the HypoPINN. We minimise (\ref{eq:3}) along with (\ref{eq:2}) as the log-likelihood term and Gaussian prior (Tikhonov regularisation). We perform the minimization for $3,000$ epochs and predict the eikonal solution with the last epoch's weights, thus localizing the hypocenter location based on (\ref{eqn:source_location}).

\begin{figure*}[!htb]
  \centering
  \includegraphics[width=\textwidth]{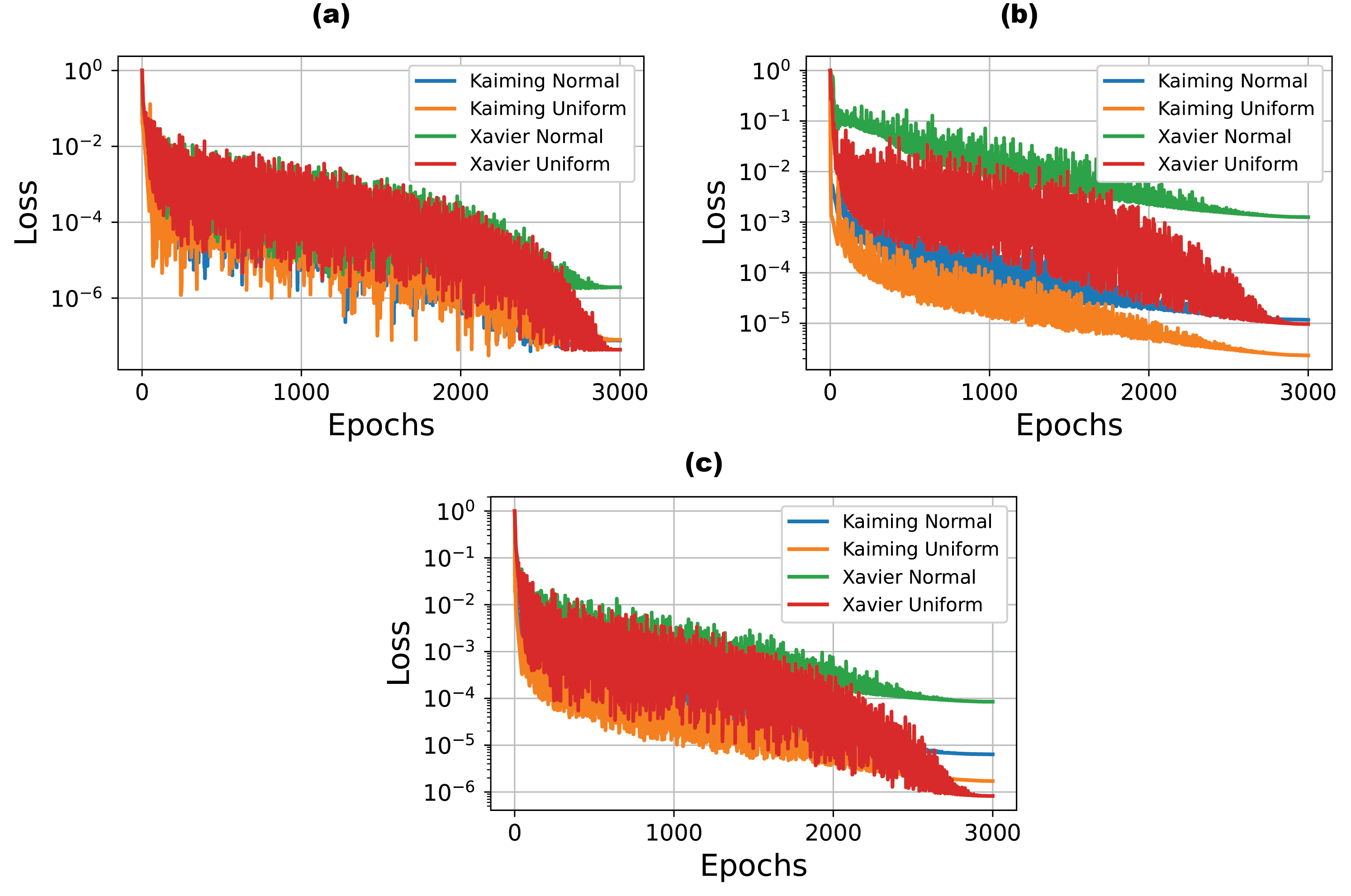}
  \caption{Random weights initializations example: Convergence history of (a) data loss, (b) PDE loss, (c) total loss.}
  \label{fig:2}
\end{figure*}

\begin{figure*}[!htb]
  \centering
  \includegraphics[width=\textwidth]{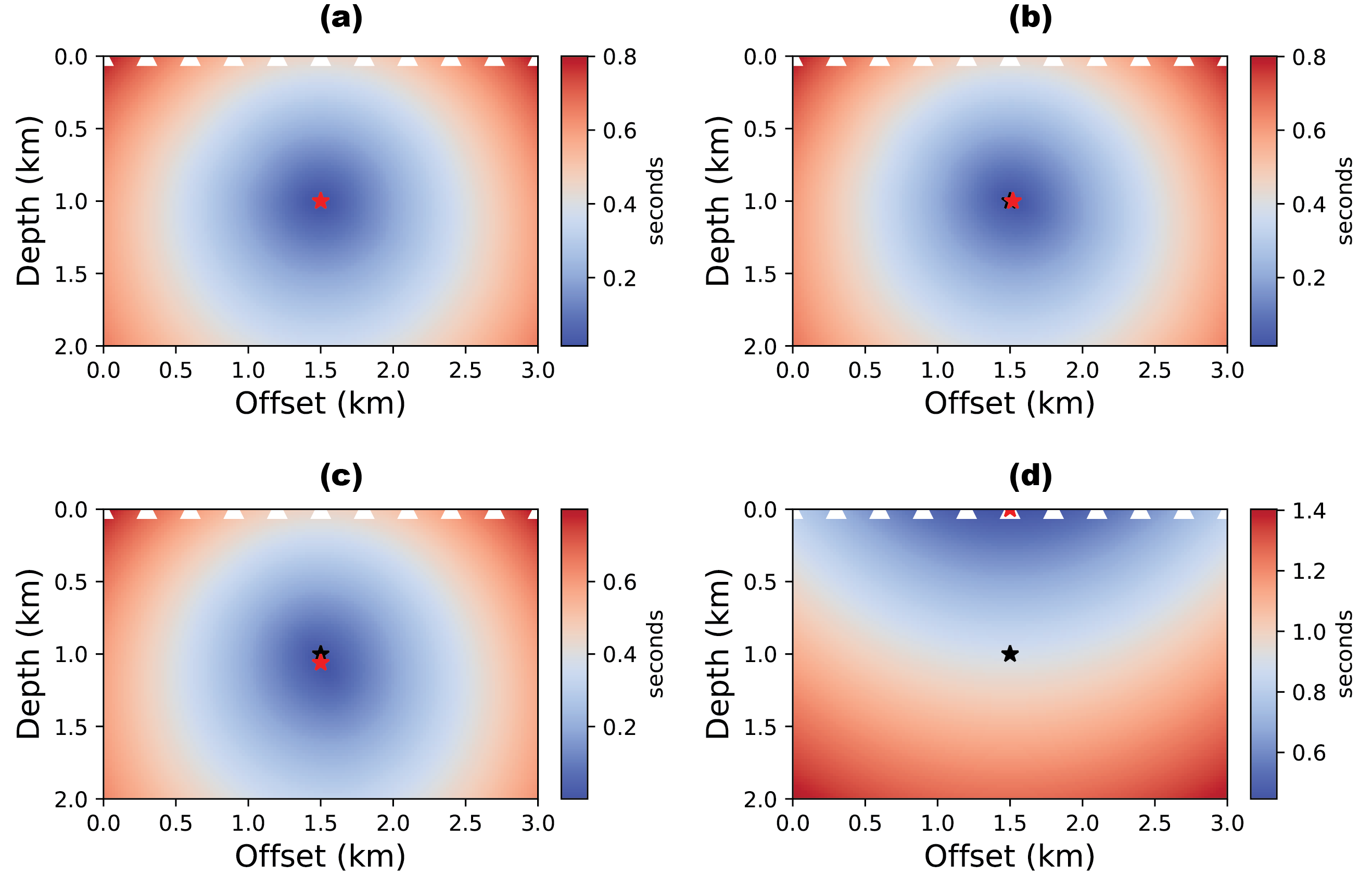}
  \caption{Random weights initializations example: Reconstructed eikonal solution for four different initial weights initialization sampled from (a) Kaiming normal distribution, (b) Kaiming uniform distribution, (c) Xavier normal distribution, and (d) Xavier uniform distribution. There are two stars depicting the true and estimated locations of the microseismic event, and in some cases they overlap.}
  \label{fig:3}
\end{figure*}

In Fig. \ref{fig:3}, we observe that the microseismic localizations from the eikonal solutions visually match the true location shown in Fig. \ref{fig:1}(b) for all distributions except for the Xavier uniform prior. Its solution is biased towards observed data on the surface. We also localize the hypocenter location by taking the location of the eikonal solutions where its value is minimum, and relatively accurate hypocenter locations are estimated for the three previously mentioned prior distributions as denoted by the red star in the same figure. Based on this experiment, we observed the effect of random weights initialization on the HypoPINN prediction as the eikonal solutions and hypocenter locations vary significantly with different initial weights realization. For example, the losses history in Fig. \ref{fig:2} for all the initial weights are showing convergence. Although the initial weights sampled from Xavier uniform distribution recorded the lowest total loss value, it failed to predict the eikonal solution correctly, thus localizing the hypocenter at the wrong location. This signifies the ill-posedness of this problem; HypoPINN with such prior weight distribution leads the training process to a bad local minimum, which belongs to the null space of the HypoPINN's solution space despite having the lowest total loss value.


\begin{figure*}[!htb]
  \centering
  \includegraphics[width=\textwidth]{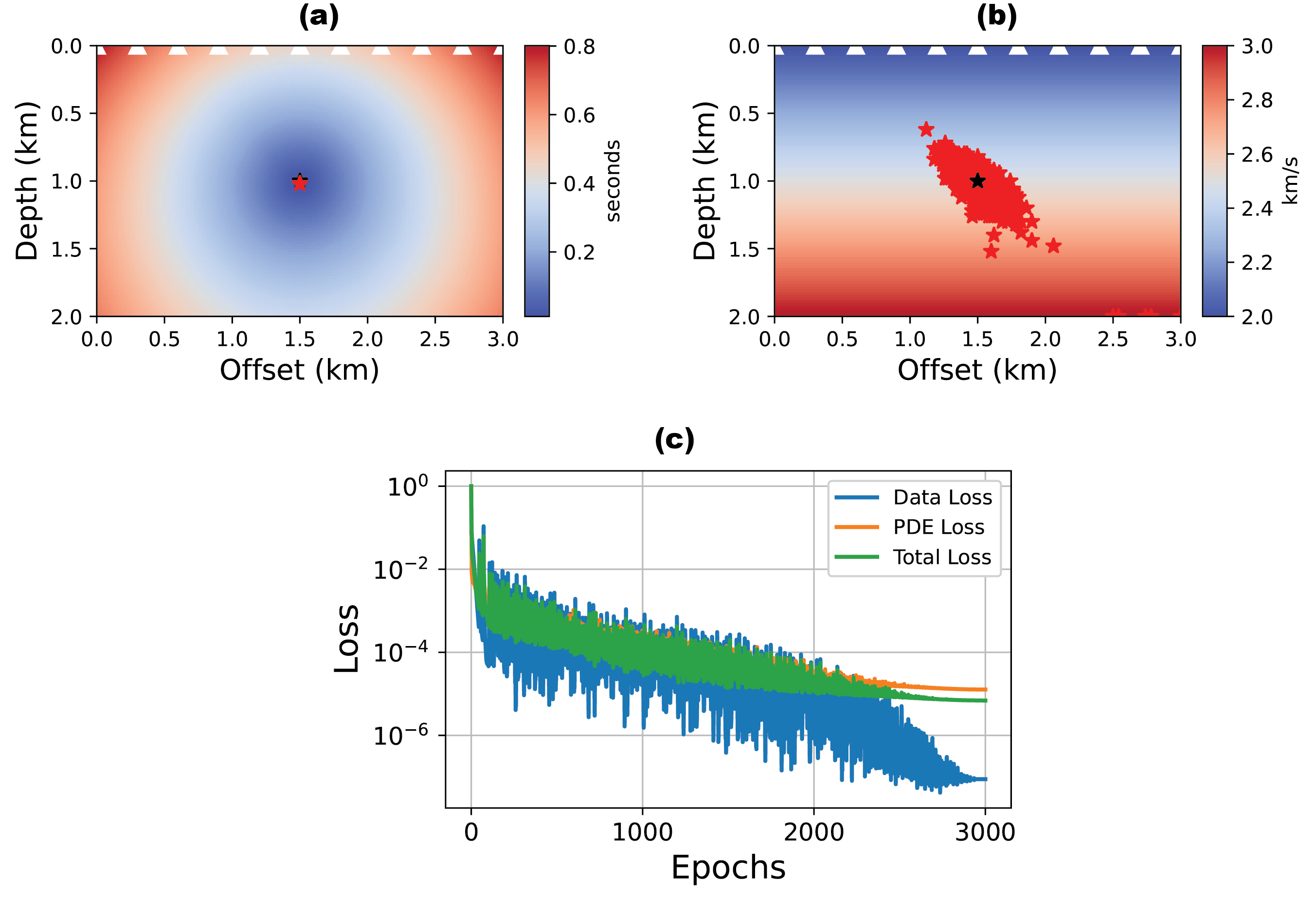}
  \caption{Predictive uncertainty of the hypocenter location associated with weights' realizations $\mathbf{\theta}$ from the Laplace approximation for vertically varying velocity model. (a) The reconstructed eikonal solution. The black and red stars denote the true source and the estimated $\mathbf{\theta}_{MAP}$ hypocenter locations, respectively. (b) The locations of hypocenter associated with 1000 $\mathbf{\theta}$ realizations from the Laplace approximation denoted by red stars. (c) Convergence history of HypoPINN's loss.}
  \label{fig:4}
\end{figure*}

Next, we repeat the experiment to study the HypoPINN's predictive uncertainties based on the Laplace approximation. Based on previous results, we consider Kaiming normal distribution as the prior distribution, and we initialize our initial weights by sampling from it. Similarly, in Fig. \ref{fig:4}(a), we observe that the eikonal solution and its hypocenter location visually matched the analytical one. Based on this result, we take the last epoch's weights as our $\mathbf{\theta}_{MAP}$ for the uncertainty analysis using the Laplace approximation.

To study the uncertainties propagations from the HypoPINN's weights to the predictive solution, we construct the Laplace approximation to the posterior distribution as in (\ref{eq:5}). This work considers the diagonal approximation of the covariance matrix described in the previous section with $2,737$ learnable network parameters. With this approximation, we sample $1000$ weights' realizations and perform the eikonal solution predictions by realizing HypoPINN with those respective weights. Based on those realizations, we have $1000$ predicted eikonal solutions, and we obtain the predicted hypocenter locations from those solutions. The results for this predictive uncertainty are illustrated in Fig. \ref{fig:4}(b). We observe that the predictions of the eikonal solution and the hypocenter locations vary significantly with different HypoPINN's weights realization, as shown in the previous example. This shows that the uncertainty in the HypoPINN's weights propagates into the predictive solution and significantly influences the prediction. In Fig. \ref{fig:4}(b), we also observe the uncertainty of the hypocenter is larger in the depth direction as we use surface recordings. However, the predictive uncertainty depends on the loss landscape of the hypoPINN. A different loss landscape where the MAP solution lands will give a significantly different predictive uncertainty. For example, we could observe that several predicted hypocenter locations are completely far from the true location and no longer reflect the physical constraint. This shows us that the predictive uncertainty is sensitive to the hypoPINN's loss landscape. Interested readers may refer to \cite{li2017} and \cite{garipov2018loss} to learn more about the loss landscape of neural networks and its effects on the network's predictive uncertainties. The examples shown here highlight critical features of this predictive uncertainty on the seismic hypocenter location.

\subsection{Otway velocity model}
\begin{figure}[!htb]
  \centering
  \includegraphics[width=0.5\textwidth]{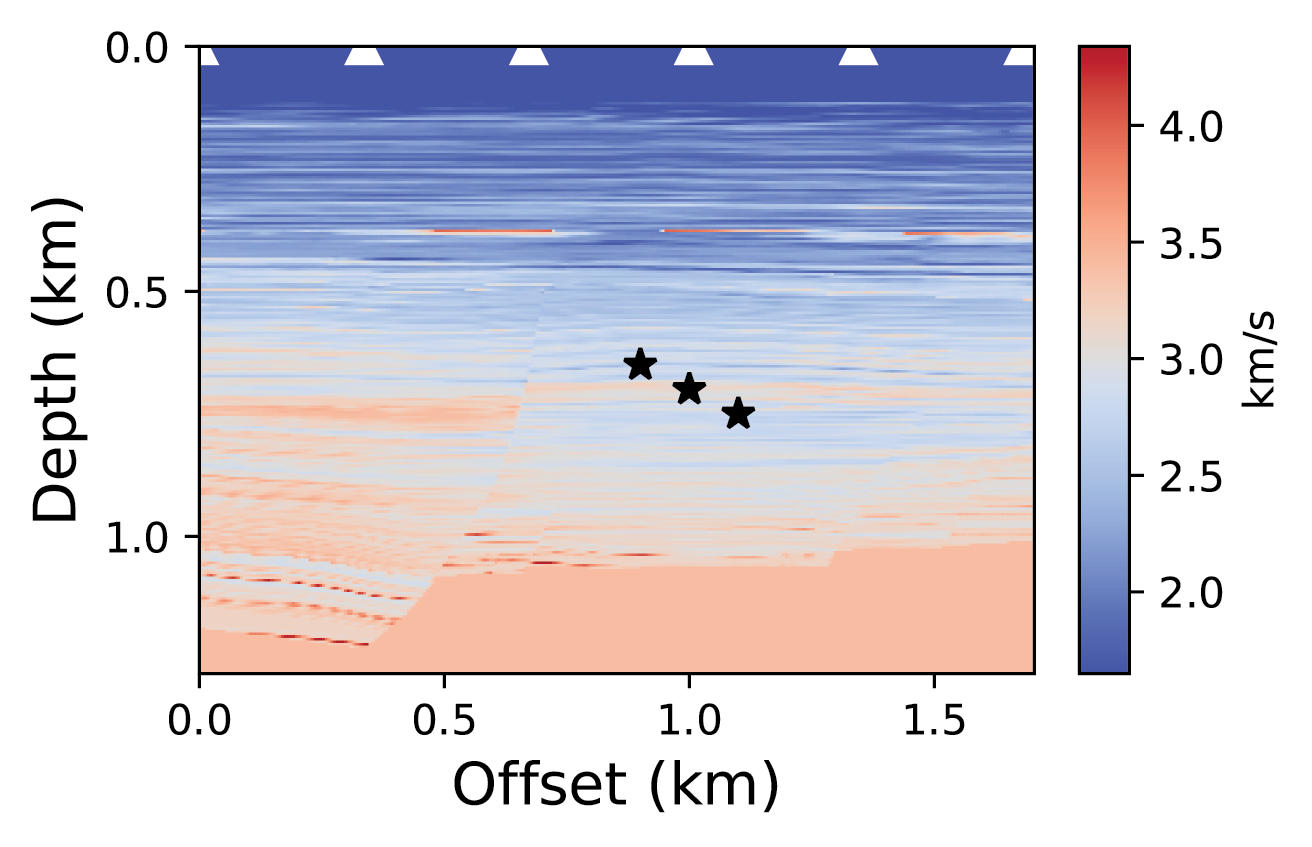}
  \caption{The Otway velocity model for hypocenter locations estimation. The black stars denote three different true source locations.}
  \label{fig:5}
\end{figure}

This example considers the Otway velocity model which belongs to Stage 2C of the Otway project by CO2CRC Limited in Australia with $1.28$ km depth and extended laterally for $1.70$ km \cite{glubokovskikh2016seismic}. Multiple point-source locations at depth and lateral positions ranging from ($0.7$ km, $1.0$ km), ($0.75$ km, $1.1$ km), and ($0.65$ km, $0.9$ km) are considered, respectively. The model is illustrated in Fig. \ref{fig:5} with the black stars depicting the point-source locations. The model is discretised on a $246 \times 305$ grid with a grid spacing of $8$ m along both axes. We use the Fast Marching Method \cite{sethian1996fm} to compute the eikonal solutions for each source location. The observed data are obtained from $6$ regularly spaced receivers at the surface of the model. 

\begin{figure*}[!t]
  \centering
  \includegraphics[width=\textwidth]{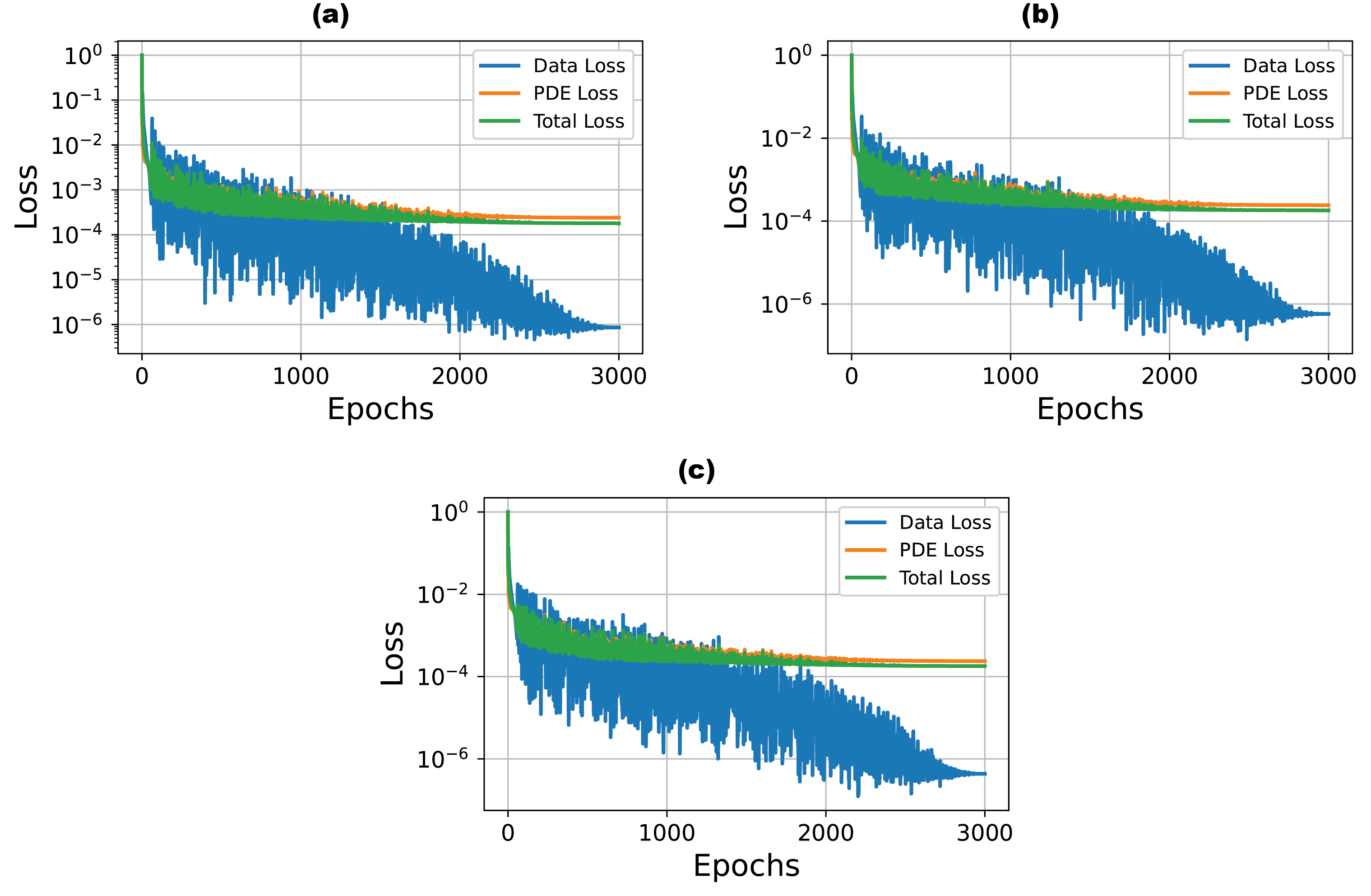}
  \caption{The convergence history of HypoPINN's training loss for three different source locations. HypoPINN is trained independently for each source location. (a) First source location at $0.7$ km depth and $1.0$ km lateral. (b) Second source location is at $0.75$ km depth and $1.1$ km lateral. (c) Third source location at $0.65$ km depth and $0.9$ km lateral.}
  \label{fig:6}
\end{figure*}

\begin{figure*}[!htb]
  \centering
  \includegraphics[width=\textwidth]{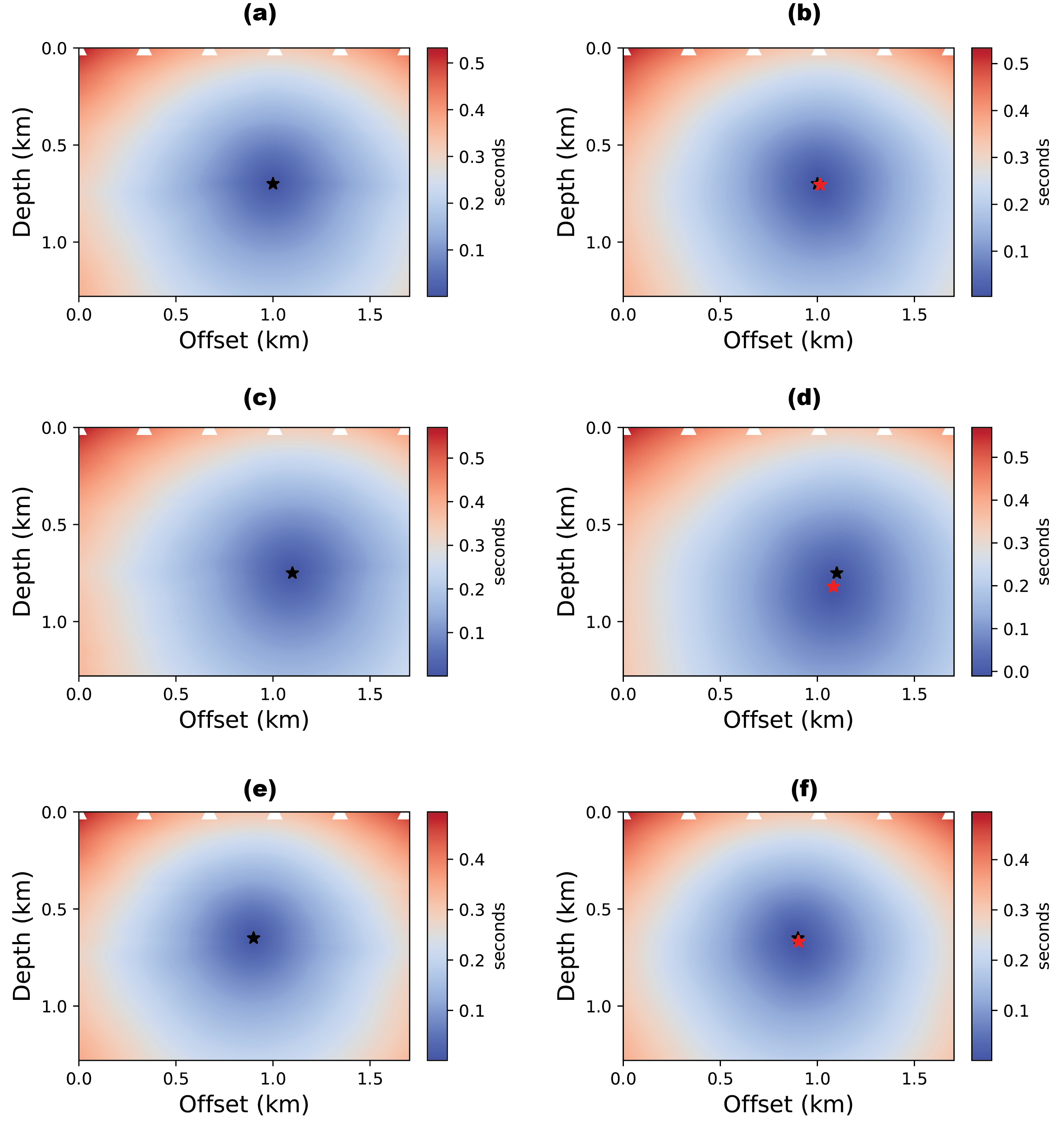}
  \caption{Reconstructed eikonal solution for three different source locations for the Otway velocity model. (a)(c)(d) True eikonal solution obtained through fast marching method, and (b)(d)(e) reconstructed eikonal solution using HypoPINN for source locations at depth and lateral positions of ($0.7$ km, $1.0$ km), ($0.75$ km, $1.1$ km), and ($0.65$ km, $0.9$ km), respectively. The black and red stars denote the true and estimated hypocenter locations, respectively.}
  \label{fig:7}
\end{figure*}

For each source location in the model, an independent HypoPINN network with $2,737$ learnable network parameters is trained to estimate the hypocenter locations. 
The initial HypoPINN's weights are initialized by sampling from the Kaiming normal distribution. We randomly sample $5000$ points in the domain for training the HypoPINN. Similar to the previous example, we minimise (\ref{eq:3}) with (\ref{eq:2}) as the log-likelihood term and the Gaussian prior (Tikhonov regularisation). We perform the minimisation for $3,000$ epochs and consider the last epoch's weights as $\mathbf{\theta}_{MAP}$ for the eikonal solution prediction, thus, estimating the hypocenter locations. The convergence of HypoPINN's training for all three scenarios is shown in Fig. \ref{fig:6}. Also, in Fig. \ref{fig:7}, overall predictions of eikonal solutions and the hypocenter locations are relatively accurate except for the slight deviation observed in Fig. \ref{fig:7}(d) for the second scenario. 

\begin{figure*}[!htb]
  \centering
  \includegraphics[width=\textwidth]{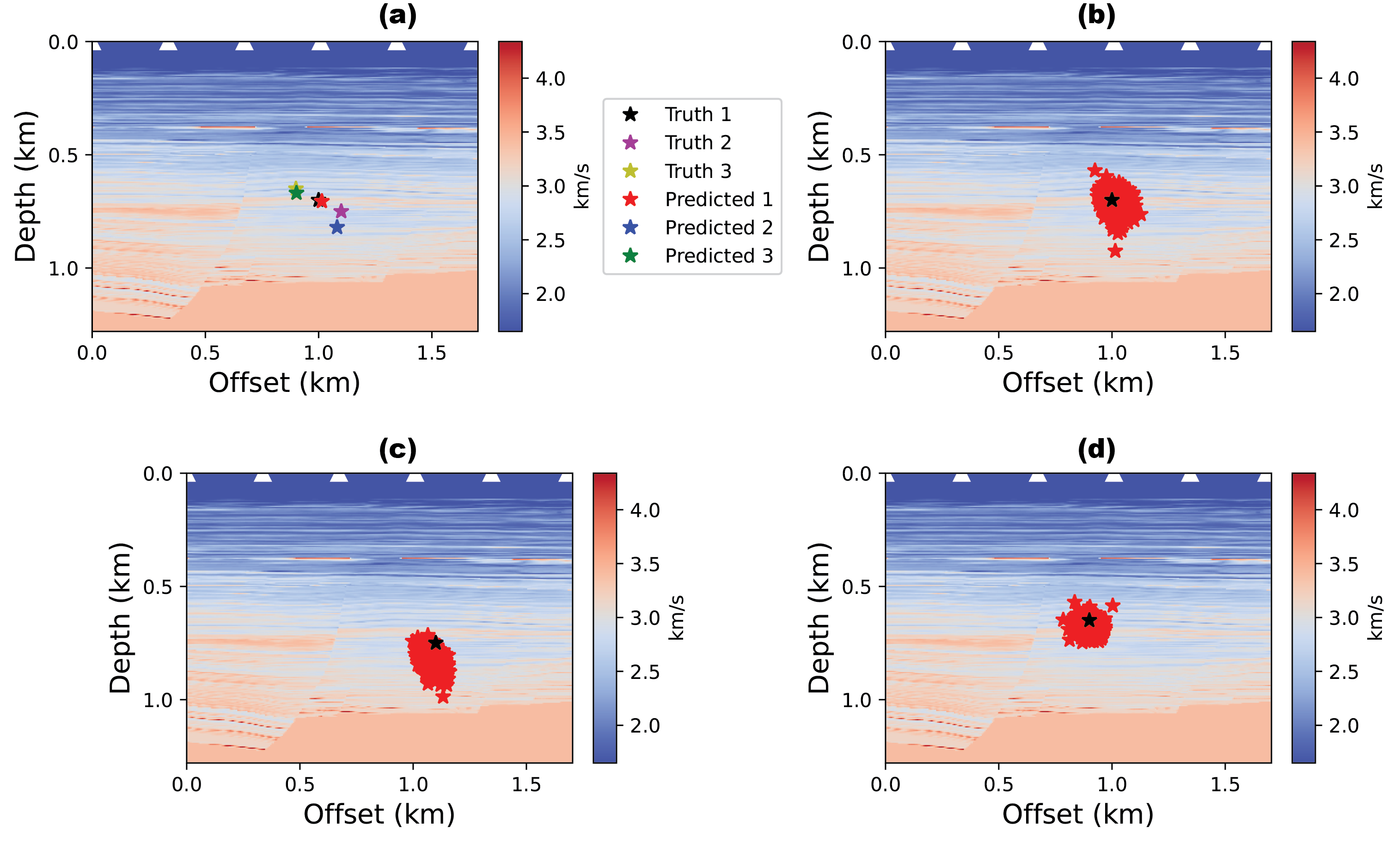}
  \caption{Predictive uncertainty of the hypocenter location associated with weights' realizations $\mathbf{\theta}$ from Laplace approximation for Otway velocity model. (a) The estimated $\mathbf{\theta}_{MAP}$ hypocenter locations in comparison to the true ones for three different source locations. (b)(c)(d) The locations of hypocenter associated with 1000 $\mathbf{\theta}$ realizations from the Laplace approximation denoted by red stars for source locations at depth and lateral positions of ($0.7$ km, $1.0$ km), ($0.75$ km, $1.1$ km), and ($0.65$ km, $0.9$ km), respectively. }
  \label{fig:8}
\end{figure*}

We proceed with the predictive uncertainties study by constructing the Laplace approximation to the posterior distribution as in (\ref{eq:5}) independently for each scenario in this example. Similar to the previous example, we consider the diagonal approximation of the covariance matrix for the Laplace approximation around the $\mathbf{\theta}_{MAP}$. We sample $1000$ weights' realizations and perform the eikonal solution predictions by realizing HypoPINN with those weights for each scenario. For the first and second scenarios in Fig. \ref{fig:8}(b) and (c), we observe the uncertainty of the hypocenter is larger in the depth direction. Yet, in the second scenario, the uncertainty of the hypocenter is biased towards deeper depth due to the inaccurate prediction of its eikonal solution and hypocenter location. In the third scenario, as depicted in Fig. \ref{fig:8}(d), we observe the predictive uncertainty for the hypocenter locations is well-constrained in depth and the lateral direction. As discussed earlier in the previous example, the predictive uncertainty of HypoPINN depends on the loss landscape surrounding the MAP solution location. This example demonstrates this phenomenon through three independent HypoPINN training for three different source locations. Nevertheless, with a limited number of observed data available and the usage of surface recordings, HypoPINN can recover relatively accurate prediction of the eikonal solution and its hypocenter locations with well-defined predictive uncertainty.


%% file: Sections/05_Discussion.tex
\section{Discussion}
This work focuses on developing an inversion framework for hypocenter localizations directly using PINNs and investigating its predictive uncertainties or, simply, forward modeling uncertainties in the context of HypoPINN. This predictive uncertainty is different from the physical model uncertainty, in which the quantity of interest is the physical quantity, e.g., hypocenter locations, velocity, etc. Our quantity of interest here is the PINN's network parameters $\mathbf{\theta}$. In the numerical examples, we demonstrated the efficacy of the proposed methodology in obtaining robust hypocenter location. However, localizing hypocenter locations alone is insufficient as many factors influence the predictive uncertainty of a neural network; thus, representing uncertainty and understanding its source is crucial for decision-making. This discussion will generally focus on the challenges and limitations of HypoPINN in localizing hypocenter locations and Laplace approximation in representing its predictive uncertainties.

\subsection{Factored vs. unfactored eikonal equation for HypoPINN}
Solving the eikonal equation using PINN was first introduced by \cite{Waheed2021} where the authors leveraged the factored eikonal equation \cite{fomel2009factored} to avoid the singularity due to the point source. Formulating the problem as such constrains the null space and regulates the ill-posedness of a neural network as it only requires learning a scaling factor. Also, the factored eikonal formulation provides a priori information regarding the geometry of the traveltime as an assistant to the assigned neural network to learn and solve the eikonal equation accurately. However, this approach requires the source locations to be known, limiting its applicability in solving the hypocenter location problem. 

HypoPINN is leveraging the unfactored or original eikonal equation in solving the hypocenter locations estimation. The main challenge for PINN in solving an eikonal equation and thus estimating hypocenter locations is the need to learn the eikonal equation as a whole by optimizing a large unconstrained solution space due to no prior information in assisting its learning. This makes the neural network learning process ill-posed as it is limited by large null-space solutions. This phenomenon is observed in our first numerical example where HypoPINN failed to predict the eikonal solution for the initial weights sampled from Xavier uniform distribution despite its loss history showing convergence. Nevertheless, these challenges can be alleviated by several strategies such as designing a task-specific network architecture, different initial weights' prior distribution, optimal hyperparameters tuning, etc. However, such strategies will influence the neural network's loss landscape, thus affecting its predictive uncertainties \cite{li2017, garipov2018loss, wilson2020bayesian}.

\subsubsection{Predictive uncertainties estimation with Laplace approximation}
Laplace approximation is arguably the simplest family of approximations for the intractable posteriors of deep neural networks. It forms a Gaussian approximation to the exact posterior. Its mean equals the MAP estimate of a neural network, and its covariance equals the negative inverse Hessian (i.e., approximation thereof) of the loss functions evaluated at the MAP estimate. Due to its simplicity, the Laplace approximation can be applied to any pre-trained neural network in a cost-efficient post hoc manner. The main ingredient of the Laplace approximation is its covariance estimation, and there are various approaches to approximate the covariance \cite{Ritter2018, Daxberger2021}. A good covariance approximation is important in predicting the uncertainty embedded in the neural network training as it affects the quality of uncertainty representation. However, good covariance approximation is commonly computationally limited by the size of learnable network parameters. This work considers the diagonal approximation, which is the simplest approximation approach in studying the HypoPINN's predictive uncertainties. Thus, such representation of uncertainties might be either under-or over-estimated. Also, Laplace approximation relies on crude approximations of the posterior distribution as the posterior is intractable due to the neural networks' size and nonlinearity.

%% file: Sections/06_Conclusion.tex
\section{Conclusion}
We developed HypoPINN -- a PINN-based algorithm for hypocenter localization. We trained a neural network by minimizing a loss function formed by the misfit of the observed and predicted traveltimes and the eikonal residual term. Given picked arrival times for an event, this provides predicted traveltimes in the entire computational domain. The hypocenter location is then given by the location of the minimum travel time. We also introduced an approximate Bayesian framework for estimating predictive uncertainty or, simply, forward modeling uncertainty in the context of HypoPINN. We investigated the uncertainties propagation from the random realizations of HypoPINN's weights and biases using the Laplace approximation to the predicted solutions. This generally opens up new pathways in investigating the training dynamics of PINN, especially in the PINN's weights initialization to obtain a correct and robust solution. Through numerical examples, we demonstrated the efficacy of the proposed methodology in obtaining robust hypocenter locations equipped with predictive uncertainty estimation, even in the case where sparse traveltime observations are available. Once a neural network model is trained, we may use transfer learning to obtain new hypocentre locations in almost real-time. This allows for computationally efficient and robust hypocenter localization that may enable the goal of real-time microseismic monitoring.

%% file: Sections/07_Acknowledgement.tex
\section*{Acknowledgment}
We thank Matteo Ravasi from KAUST for fruitful discussions and constructive suggestions for this work. This publication is based on work supported by King Abdullah University of Science and Technology (KAUST).

\section*{Code Availability}
All accompanying codes will be publicly available at https://github.com/izzatum/laplace-hypopinn.